\documentclass[conference]{IEEEtran}
\IEEEoverridecommandlockouts
\usepackage{cite}
\usepackage{multirow,makecell}
\usepackage{amsmath,amssymb,amsfonts}
\usepackage{algorithmic}
\usepackage{graphicx}
\usepackage{textcomp}
\usepackage{xcolor}
\def\BibTeX{{\rm B\kern-.05em{\sc i\kern-.025em b}\kern-.08em
    T\kern-.1667em\lower.7ex\hbox{E}\kern-.125emX}}
\begin{document}

\title{Prediction of COVID-19 Patients' Emergency Room Revisit using Multi-Source Transfer Learning\\
\thanks{*Yuelyu Ji, Yuhe Gao, and Runxue Bao contributed equally. Corresponding author: Ye Ye}
}

\author{
\IEEEauthorblockN{Yuelyu Ji* }
\IEEEauthorblockA{\textit{Department of Information Science} \\
\textit{School of Computing and Information} \\
\textit{University of Pittsburgh}\\
Pittsburgh,USA\\
yueluji@gmail.com}
\and

\IEEEauthorblockN{Yuhe Gao* }
\IEEEauthorblockA{\textit{Department of Biomedical Informatics} \\
\textit{School of Medicine} \\
\textit{University of Pittsburgh}\\
Pittsburgh, USA\\
yug51@pitt.edu}
\and

\IEEEauthorblockN{Runxue Bao*}
\IEEEauthorblockA{\textit{Department of Electrical and}\\\textit{ Computer Engineering} \\
\textit{Swanson School of Engineering} \\
\textit{University of Pittsburgh}\\
Pittsburgh, USA \\
baorunxue@gmail.com}
\and

\IEEEauthorblockN{Qi Li}
\IEEEauthorblockA{\textit{School of Business} \\
\textit{State University of New York at New Paltz}\\
New Paltz, USA \\
liq11@newpaltz.edu}
\and

\IEEEauthorblockN{Disheng Liu}
\IEEEauthorblockA{\textit{Department of Information Science} \\
\textit{School of Computing and Information} \\
\textit{University of Pittsburgh}\\
Pittsburgh, USA \\
dsliu0353@gmail.com}
\and

\IEEEauthorblockN{Yiming Sun}
\IEEEauthorblockA{\textit{Department of Electrical and} \\\textit{ Computer Engineering} \\
\textit{Swanson School of Engineering} \\
\textit{University of Pittsburgh}\\
Pittsburgh, USA \\
yis108@pitt.edu}
\and

\IEEEauthorblockN{Ye Ye}
\IEEEauthorblockA{\textit{Department of Biomedical Informatics} \\
\textit{School of Medicine} \\
\textit{University of Pittsburgh}\\
Pittsburgh, USA \\
yey5@pitt.edu}

}
\maketitle

\begin{abstract}
The coronavirus disease 2019 (COVID-19) has led to a global pandemic of significant severity. In addition to its high level of contagiousness, COVID-19 can have a heterogeneous clinical course, ranging from asymptomatic carriers to severe and potentially life-threatening health complications. Many patients have to revisit the emergency room (ER) within a short time after discharge, which significantly increases the workload for medical staff. Early identification of such patients is crucial for helping physicians focus on treating life-threatening cases. In this study, we obtained Electronic Health Records (EHRs) of 3,210 encounters from 13 affiliated ERs within the University of Pittsburgh Medical Center between March 2020 and January 2021. We leveraged a Natural Language Processing technique, ScispaCy, to extract clinical concepts and used the 1001 most frequent concepts to develop 7-day revisit models for COVID-19 patients in ERs. The research data we collected were obtained from 13 ERs, which may have distributional differences that could affect the model development. To address this issue, we employed a classic deep transfer learning method called the Domain Adversarial Neural Network (DANN) and evaluated different modeling strategies, including the Multi-DANN algorithm (which considers the source differences), the Single-DANN algorithm (which doesn't consider the source differences), and three baseline methods: using only source data, using only target data, and using a mixture of source and target data. Results showed that the Multi-DANN models outperformed the Single-DANN models and baseline models in predicting revisits of COVID-19 patients to the ER within 7 days after discharge (median AUROC = 0.8 vs. 0.5). Notably, the Multi-DANN strategy effectively addressed the heterogeneity among multiple source domains and improved the adaptation of source data to the target domain. Moreover, the high performance of Multi-DANN models indicates that EHRs are informative for developing a prediction model to identify COVID-19 patients who are very likely to revisit an ER within 7 days after discharge.

\end{abstract}

\begin{IEEEkeywords}
deep transfer learning, domain adversarial neural network, multiple sources, COVID-19, emergency room revisit 
\end{IEEEkeywords}

\section{INTRODUCTION}
The coronavirus disease 2019 (COVID-19) has led to a global pandemic of significant severity, characterized by rapid spread among individuals and higher morbidity and mortality rates than previous respiratory infectious diseases (e.g., influenza), particularly among older adults and individuals with pre-existing health conditions\cite{r2}. This disease can lead to acute respiratory failure and other related illnesses (e.g., multiorgan dysfunction), which may cause massive hospital readmission and emergency room (ER) revisits\cite{r3,peiris2022hospital}.







It is critical to reduce COVID-19 patients' ER revisits. First, a revisit within a short period suggests inadequate treatment and lack of timely follow-up. Identifying and providing timely outpatient appointments for high-risk patients, who are likely to revisit ERs, will lead to better treatments and prognoses. Second, during the COVID-19 pandemic, ERs were already overloaded with new patients. Readmission of recently discharged COVID-19 patients exacerbates this burden. Reducing these revisits would help physicians focus on treating other life-threatening cases, such as heart attacks and trauma. Moreover, reducing the number of contagious patients in ERs would also ease the management burden of hospital infection control and prevention. 

Unfortunately, it is not easy for ER clinicians to predict which group of patients are more likely to revisit ERs very soon, because COVID-19 patients can have a heterogeneous clinical course, ranging from asymptomatic carriers to multiorgan failure \cite{peiris2022hospital}. A risk assessment model would be very useful, and it might be developed using machine learning and Electronic Health Records (EHRs). Developing a risk assessment model often needs abundant training samples with class labels. When training data in one ER is insufficient, leveraging data from other ERs could be necessary.






Transfer learning, a machine learning method that facilitates the sharing of data from one or multiple source domain(s) to a new target domain, may greatly reduce the resource requirements and computational cost \cite{pan2010survey}. In recent years, deep transfer learning, transfer learning techniques that develop deep neural network models, has achieved great success in computer vision \cite{zhao2020review}, natural language processing \cite{ruder2019transfer}, and many biomedical areas, including image analysis \cite{balachandar2020accounting, muhammad2020deep,alzubaidi2021role}, bioinformatics \cite{zhao2020multi, kim2021anticancer}, EHR natural language processing \cite{sachan2018effective, lin2020does}, drug discovery \cite{dana2018deep}, and mobile health \cite{saeedi2020collaborative}. Specifically, a classic deep transfer learning algorithm, Domain Adversarial Neural Network (DANN) \cite{ganin2016domain}, has been successfully applied in biomedical image processing \cite{brion2021domain, lafarge2017domain} and infectious disease detection \cite{ye2021deep}.

In this study, we applied the DANN algorithm to the COVID-19 ER revisit prediction task in single-source and multi-source scenarios, and compared them with baseline strategies. The contributions of this study are twofold: 

\begin{itemize}
\item We successfully leveraged historical EHRs to develop prediction models of ER revisits for COVID-19 patients. These models will be able to help ER clinicians identify high-risk patients so that clinicians can allocate more medical resources to those patients to achieve better health outcomes. 
\item With real-world EHRs, we compared different modeling strategies (two deep transfer learning methods with baseline methods) and found that the multi-source transfer learning strategy can handle the heterogeneity among multiple source domains well. 
\end{itemize}
\section{RESEARCH DATA}
This study focused on active COVID-19 infections, which are defined by ICD10 codes, specifically U07.1 (COVID-19) and J12.82 (Pneumonia due to COVID-19). 
With historical EHRs from 13 affiliated ERs within the University of Pittsburgh Medical Center (UPMC), a revisit is defined as any ER encounter (occurring at any of the 13 ERs) within 7 days after the date of a previous ER encounter and the revisit still had a diagnosis of COVID-19. The revisits are positive cases in our research data. Negative cases correspond to patients who 
\begin{itemize}
    \item were diagnosed with COVID-19 but did not revisit any of the 13 ERs within 7 days.
    \item were diagnosed with COVID-19 and revisited one of the 13 ERs within 7 days, but the revisit was not related to a COVID-19 diagnosis.
\end{itemize}

Research data include 3,210 ER encounters (to 13 UPMC ERs) between March 16, 2020, and  January 31, 2021. 
The frequency counts are available in Table \ref{tab3}. The number of COVID-19 encounters ranges from 83 to 594. The revisit rate ranges from 16\% (ER 5) to 35\% (ER 13), with an overall revisit rate of 22\% (708 over 3,210). The R3 team at the University of Pittsburgh retrieved encounter information and anonymized discharge notes from the Neptune Research Data Warehouse\cite{r20}. The project is approved by the IRB of the University of Pittsburgh (STUDY19050197).

For each discharge note, we applied ScispaCy \cite{r21} to identify medical concepts
and code them with UMLS Concept Unique Identifier (CUIs). The most frequent 1,001 CUIs in all records have been selected to represent the common features. For each encounter, each CUI has three labels: Present (P), identified in notes and not negated; Negated (N), identified in notes but negated for all its appearances; Missing (M), not identified. We used these features and the class label of revisit or not to develop revisit risk models: 1 for an ER encounter that had at least one COVID-19 related revisit within 7 days of discharge; 0 for no COVID-19 related revisit. 

\begin{table*}[htbp]
\caption{THE COUNTS OF COVID-19 ENCOUNTERS TO 13 ERs IN RESEARCH DATA}
\begin{center}
\renewcommand{\arraystretch}{1.3}
\begin{tabular}{|c|c|c|c|}
\hline

\textbf{ER} & \makecell{\textbf{N. of Encounters (Positives)}} &\makecell{\textbf{N. of Encounters for Train (Positives)}} & \makecell{\textbf{N. of Encounters for Test (Positives)}} \\ 
\hline

1 & 594 (135) & 474 (112) & 120 (23) \\ \hline
2 & 483 (88) & 385 (70) & 98 (18)  \\ \hline
3 & 288 (58) & 229 (54) & 59 (4)  \\ \hline
4 & 251 (50) & 199 (38) & 52 (12) \\ \hline
5 & 233 (38) & 185 (19) & 48 (19) \\ \hline
6 & 229 (49) & 182 (40) & 47 (9) \\ \hline
7 & 216 (67) & 171 (49) & 45 (18)\\ \hline
8 & 215 (24) & 171 (11) & 44 (13) \\ \hline
9 & 183 (57) & 145 (42) & 38 (15) \\ \hline
10 & 165 (36) & 130 (25) & 35 (11) \\ \hline
11 & 149 (45) & 118 (33) & 31(12) \\ \hline
12 & 121 (32) & 95 (25) & 26 (7) \\\hline
13 & 83 (29) & 65 (23) & 18 (6) \\\hline
Total & 3210 (708) & 2549 (541) & 661 (167)  \\\hline
\end{tabular}
\label{tab3}
\end{center}
\end{table*}

\section{METHODS}
Transfer learning has been widely studied for years, and its effectiveness has been extensively discussed\cite{pan2010survey,day2017survey,zhuang2020comprehensive, jiang2022transferability}. Traditional machine learning is characterized by training and testing data using the same input feature space and the same data distribution. Addressing the data distribution difference problem, transfer learning aims to improve target learners' performance on the target domain by transferring the knowledge contained in different but related source domains. In our case, the target domain is one ER for which we need to predict the event of ER revisit. Our source domains are other ERs that share clinical data with the target ER. 

\begin{figure*}[htbp]
\centerline{\includegraphics[width=130mm]{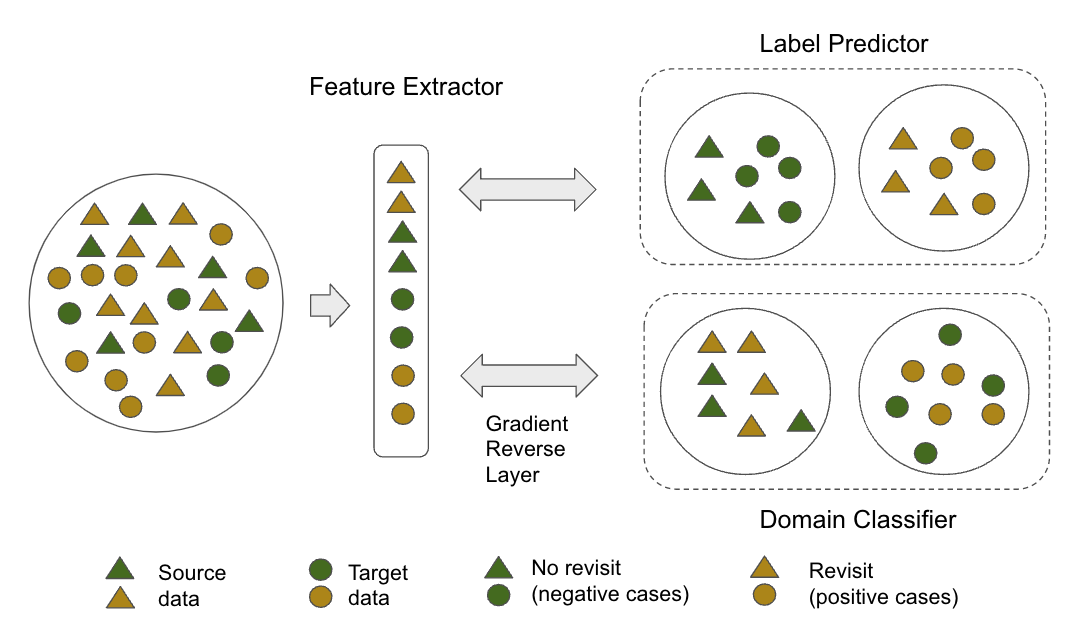}}
\caption{Simple Illustration of DANN}
\label{DANN}
\end{figure*}

The DANN algorithm \cite{ganin2016domain} is a powerful transfer learning method that makes use of the adversarial learning strategy to share knowledge from the source domain(s) to a target domain. The main goal is to learn a domain-invariant feature representation that is discriminative for the main classification task while being indiscriminate for the domain classifier. With this goal, a DANN model's architecture consists of a label predictor for the main learning task and a domain classifier for learning domain-invariant representation (Figure \ref{DANN}). A gradient reversal layer makes the model trainable through the standard back-propagation technique. In this way, the DANN algorithm facilitates the development of a common feature projection for both source domain(s) and a target domain, and this common projection is good for machine learning tasks in the target domain when using data from both source domain(s) and a target domain.

After sharing data from multiple source ERs with the target ER, there are two training strategies to handle them. One strategy is to mix data from multiple source ERs into a single data pool for training and use the DANN algorithm; this strategy is named the Single-DANN algorithm, which generates the Single-DANN models in this paper. The other strategy is to differentiate among various source ERs and treat each ER as an individual source when using the DANN algorithm, which is named the Multi-DANN algorithm, generating Multi-DANN models.

Because the Single-DANN algorithm is a special case of the Multi-DANN algorithm, we first introduce the Multi-DANN algorithm. For Multi-DANN, we consider the transfer learning problem with $M$ source domains $\{D_{s_j}\}_{j = 1}^M$ and one target domain $D_t$. Please note, when $M = 1$, a multi-DANN reduces to a single-DANN. In the $j$-th source domain $D_{s_j}$, $X_{s_j} $ and $y_{s_j} \in \{0, 1\}$ denote the source data and label respectively. We used a two-layer 256-dimension fully-connected neural network as the feature extractor $G_f$ in our model where $G_f: x_i \rightarrow \mathbb{R}^p$ is a function that mapped a given sample $x_i$ (from a source domain or a target domain) into a $p$-dimensional representation. In addition, we used a two-layer 256-dimension neural network as the predictor for the main task, which consists of a function $G_y: \mathbb{R}^p \rightarrow [0, 1]^2  $ and a binary cross-entropy loss $l_y$.  Therefore, the loss for the main task, $L_y(G_f, G_y, X, y) $, can be derived as:
\begin{eqnarray}
&& L_y\left(G_f, G_y, X, y\right)  \\ 
&=& \sum_{i} l_y\left(G_y\left(G_f\left(x_i\right), y_i\right)\right) \nonumber \\ &=& 
\sum_{i}  y_i \log \frac{1}{G_y\left(G_f\left(x_i\right)\right)} + \left(1 - y_i\right) \log \frac{1}{1 - G_y\left(G_f\left(x_i\right)\right)}, \nonumber
\end{eqnarray}
which is minimized to make the learned feature to be discriminative for the main learning task.

For the domain classifier, $G_d: \mathbb{R}^p \rightarrow [0, 1]^{M+1}  $, we also use a two-layer 256-dimension fully-connected neural network. Let $d_i \in \mathbb{R}^{M + 1}$ be the  domain label. For the $i$-th example comes from domain $j$, we have $d_{i,j} = 1$ and $d_{i,k} = 0 \text{ if} \ k \neq j$ . Note $G_d(G_f(x_i))$ estimates the probabilities that the neural network assigns $x_i$ to  $M + 1$ domains, the loss $L_d(G_f, G_d, X, d) $ for the domain classifier can be calculated as follows:
\begin{equation}
L_d(G_f, G_d, X, d) = \sum_{i}\sum_{j} d_{i, j} \log \frac{1}{G_d(G_f(x_i))_j},
\end{equation}
which is maximized to encourage the features to be domain-invariant.

Finally, a Multi-DANN model can be jointly trained by an overall objective function as
\begin{equation}
L(X, y, d) = L_y\left(G_f, G_y, X, y\right) - \lambda L_d(G_f, G_d, X, d),
\end{equation}
where $\lambda$ is a hyperparameter (our study arbitrarily used 1), which controls the trade-off between the two objectives that shape the domain-invariant features during the model learning.  Minimizing the overall loss above will simultaneously minimize the loss of the label classifier and maximize the loss of the domain classifier. Therefore, the feature representation will be discriminative for the main classification task but indiscriminate among domains. 

\begin{table}[htbp]
\caption{Experiment System Summary (All models are evaluated using target ER testing data). }
\begin{center}
\renewcommand{\arraystretch}{1.5}
\begin{tabular}{|c|c|c|c|}
\hline
\textbf{Name} & \textbf{Training} & \makecell{\textbf{Differentiate} \\ \textbf{source} \\\textbf{and target}} & \makecell{\textbf{Differentiate} \\ \textbf{sources}} \\
\hline
\makecell{Multi-DANN} & \makecell{13 ERs} & Yes & Yes \\
\hline
\makecell{Single DANN} & \makecell{13 ERs} & Yes & No\\
\hline
\makecell{Source Model} & \makecell{12 Source ERs}  & No & No  \\
\hline
\makecell{Target Model} & \makecell{Target ER} & No & No  \\
\hline
\makecell{Combined Model} & \makecell{13 ERs} & No & No \\
\hline
\end{tabular}
\label{tab1}
\end{center}
\end{table}

Besides the Single-DANN algorithm and Multi-DANN algorithm, we also designed three baseline methods, which use the same two fully connected layers for feature architecture. As listed in Table \ref{tab1}, each source model for an ER was learned using data from all ERs, except the ER as training; a target model was learned using the target ER's training dataset as training; the combined model was developed using all ERs' training datasets as training. And all models for a target ER are evaluated on the same target testing dataset.

In this study, we used 3,210 records for model development, which were split into 80\% for training and 20\% for testing (as shown in Table \ref{tab3}) based on the patients' encounter date. Encounters for training occurred earlier than encounters for testing. We used the area under the receiver operating characteristic curve (AUROC) to evaluate the performance of models.

\section{RESULTS}

\begin{figure}[htbp]
\centerline{\includegraphics[width=95mm]{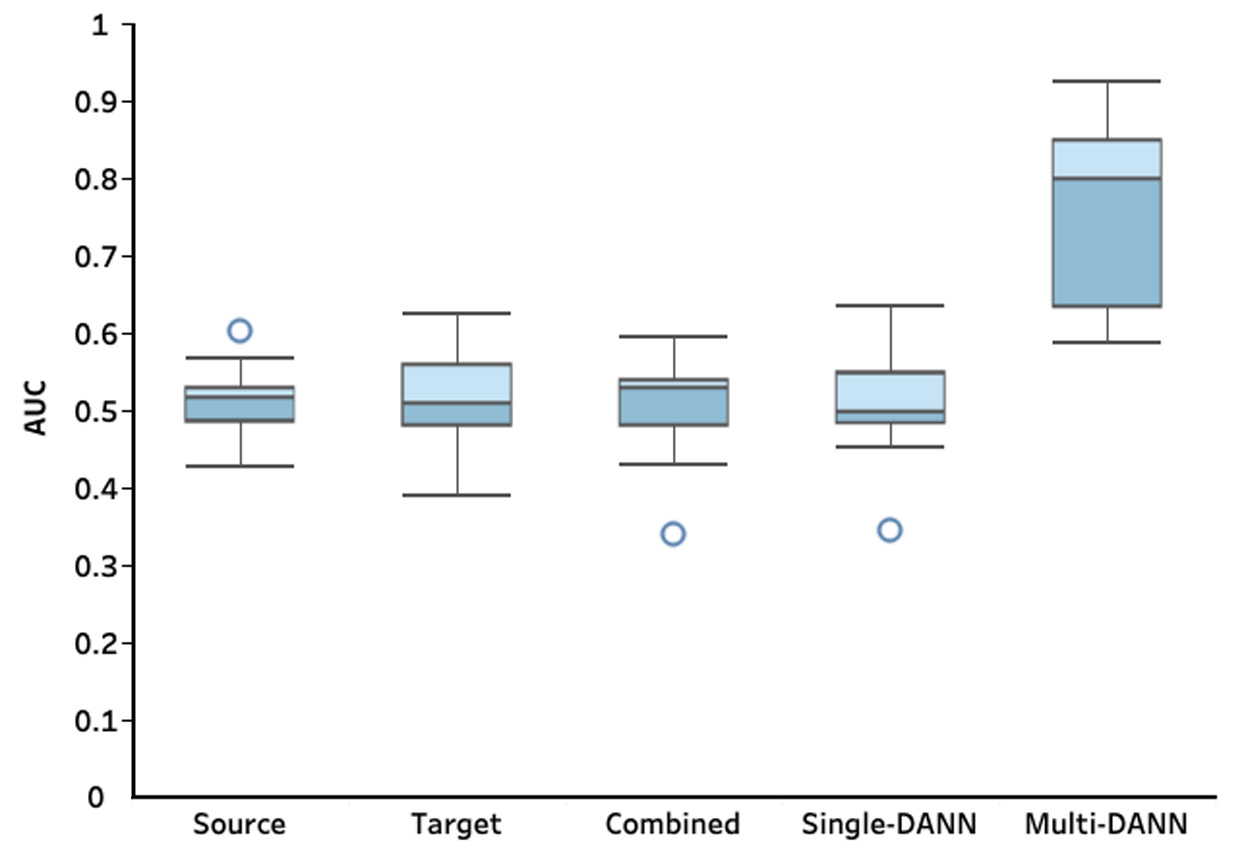}}
\caption{Boxplot of AUROCs of Models Developed with Different Strategies.}
\label{fig3}
\end{figure}

\begin{figure*}[htbp]
\centerline{\includegraphics[width=170mm]{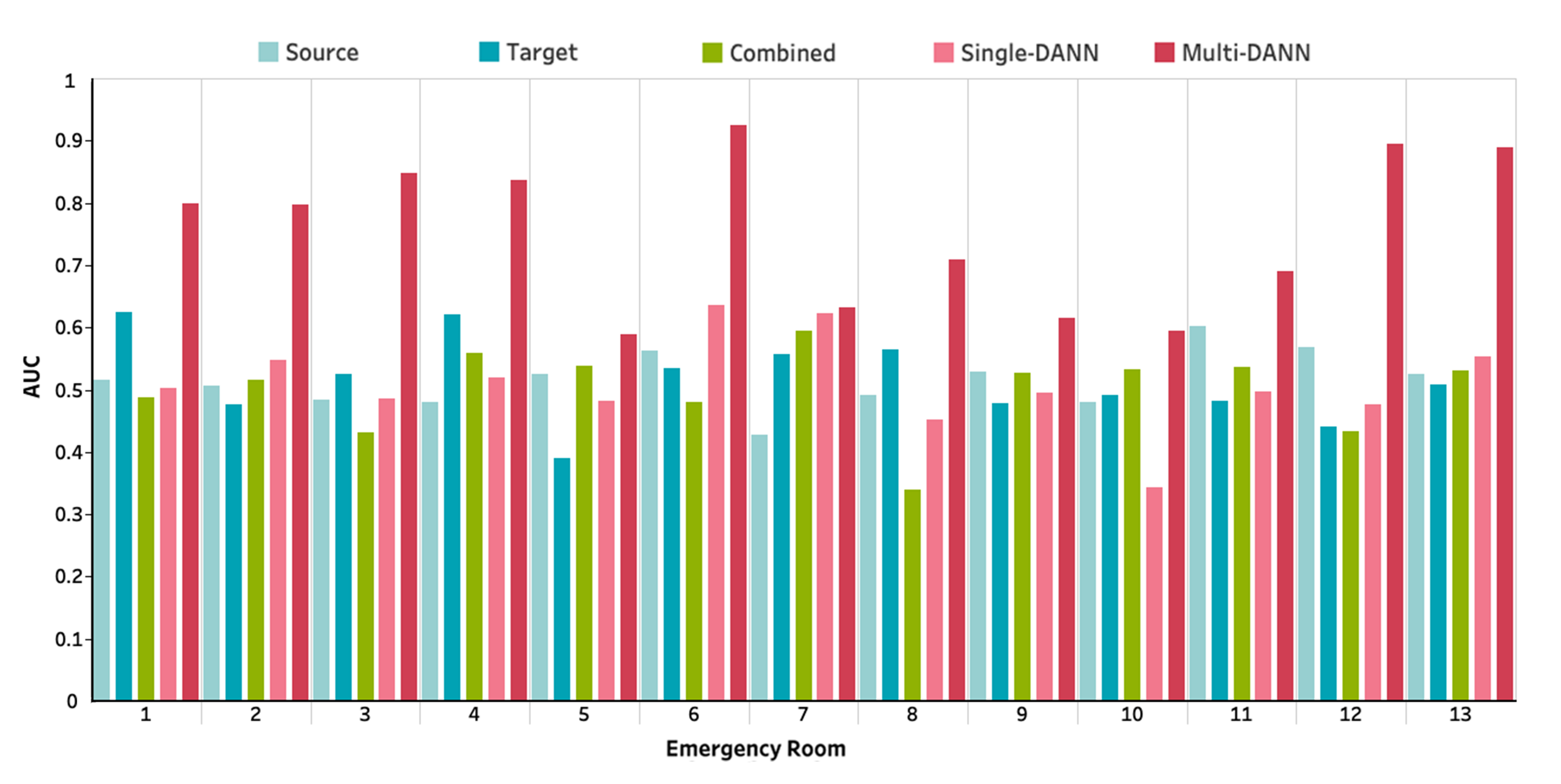}}
\caption{Comparison  of model performance for the 13 Emergency Rooms.}
\label{fig4}
\end{figure*}

We compared the performance of different models (tested in each of the 13 ERs) in Figure \ref{fig3} and Figure \ref{fig4}. As illustrated in the boxplot (Figure \ref{fig3}), the AUROC values for the baseline models (Source, Target, and Combined) are around 0.5 in the overall trend. Single-DANN models performed slightly better than baseline models, as their maximum and minimum AUROCs are slightly higher than those in all baseline models. However, the median AUROC for Single-DANN models is still close to 0.5, which indicates this modeling strategy is not reliably better. In contrast, Multi-DANN models reach the highest AUROC. Although they present a wide spread, the lowest AUROC value is still higher than the 75\% quartile of all other models. In the overall trend, Multi-DANN models performed significantly better than all other models.

Looking specifically into ERs, Figure \ref{fig4} compares the performance of different models tested in each of the 13 ERs. The results of baseline models show that either the source or target models performed well in most ERs. Specifically, for ERs 1, 4, 7, and 8, using the target ER's own training data (target baseline) resulted in the highest AUROC values among the baseline models to predict an event of ER revisit within 7 days of ER discharge; while for ERs 11 and 12, the source baseline model, which uses data from other ERs, performed better than using the target ER's own data. The combined baseline model, which uses both data from multiple source domains and target training data, did not show any significant improvement in performance. In summary, baseline models presented varying performances among ERs.

In terms of the DANN models, Multi-DANN models consistently exhibited significantly higher AUROC values across all ERs, with a median AUROC value of 0.8 and a maximum of 0.93; while Single-DANN models performed better than the baseline models in only four ER scenarios. And even in these cases, the improvements were limited, with an AUROC value not exceeding 0.65. Compared to the average performance of the baseline models, Multi-DANN models have improved predictions by 18-86\%. Furthermore, 9 out of 13 ERs' predictions have seen an improvement of over 45\% when using a Multi-DANN model as opposed to a Single-DANN model. Overall, these results suggest that Multi-DANN models performed the best, highlighting the robustness of the Multi-DANN modeling strategy.

\section{DISCUSSION}
Transfer learning can greatly facilitate the knowledge transfer from source domain(s) to a target domain. 
Using data from multiple source ERs, we applied the DANN algorithm to develop models that can predict an event of ER revisit. Results showed that Multi-DANN models can achieve significantly better performance than Single-DANN models and baseline models that do not consider the heterogeneity among multiple source domains.
Specifically, our Single-DANN models use combined data from all source ERs as a single source, and Multi-DANN models treat each individual ER as a separate source. Given Single-DANN models' low values of AUROC from our results, it can be inferred that directly combining data from different domains will not be effective in properly identifying and extracting relevant features across domains. This is in line with previous research, which has shown that compiling multiple sources of data into one source ignores the differences among sources (domain shifts) and may not be able to effectively adapt to the target domain\cite{r22, r23}. In contrast, by labeling different source data, multi-source transfer learning takes data diversity into consideration, and further reduces the impact of bias from a particular source; therefore, it can achieve domain adaptation and data augmentation in high performance  \cite{r22, r23, r24}. This is evidenced by the highest AUROC values observed in multiple target ERs in our study.

While our Multi-DANN models performed robustly across different ERs, there are several limitations in this study. First, as we are considering revisits to 13 ERs, we may miss encounters with patients who revisited other UPMC ERs and other ERs outside of the UPMC system. Second, we used the most frequent 1,001 CUIs as the common features across ER domains, which may not capture the unique characteristics of a particular ER. More influential features (e.g., hospital utilization measures) with efficient feature representations \cite{bengio2013representation, dou2022learning, dou2022sampling} will be desired to boost the performance for future real-world applications. Third, we did not explore the model architecture or hyperparameters for optimal performance, and thus, the current results may be underestimated. Fourth, we only evaluated the performance using AUROC but did not explain the developed model from a clinical perspective. Finally, all the research data used in this study are from 13 ERs located in Pittsburgh, Pennsylvania, and it is possible that the population in this location has some common features, like socioeconomic status, circulating subtypes, and local public health policy, which could constrain the generalizability of the model to other locations. 

This project successfully used EHRs to develop 7-day ER revisit models for COVID-19 patients, demonstrating the potential of using a machine-learned model to identify patients whose medical conditions might deteriorate very soon. These 
models will provide valuable probability estimations to identify high-risk patients and thus support appropriate follow-up services after ER discharges. Our future work includes developing a more reliable ER revisit model for patients with infectious respiratory diseases, using more training samples from multiple ERs and novel multi-source transfer learning techniques and large language models, such as clinical BERT \cite{alsentzer2019publicly}, BioGPT \cite{luo2022biogpt}, and chatGPT \cite{chatgpt,kung2023performance}. Future model training will go through multiple explorations on model structure and hyperparameters. 
After obtaining a successful (deep neural network) model, it is important to transform the model from a black box to a transparent box so that the developed model can provide a valid second opinion to clinicians and make a real-world impact. In addition to providing probability estimations of patients' revisit risk, it is more desirable that we define probability thresholds to reach a high recall and further develop this model into a decision support tool, which can provide actionable guidance to clinicians on various follow-up interventions (e.g., telemonitoring, app-based monitoring, virtual wards \cite{peiris2022hospital}, or outpatient visits) for patients at different risk levels.

\section{CONCLUSION}
EHRs are informative for developing a prediction model to identify COVID-19 patients who are likely to revisit the ER within 7 days. Transfer learning can facilitate data sharing from source domain(s) to a target domain. Sharing data from multiple source domains will be aided by a multi-source transfer learning technique that considers the heterogeneity among source domains, and the multi-DANN algorithm used in this study may work very well.


\section*{ACKNOWLEDGEMENT}

This publication was supported by the research grant R00LM013383 from the National Library of Medicine, National Institutes of Health. The research data were retrieved by R3, which was supported by the National Institutes of Health through Grant Number UL1 TR001857.

\bibliographystyle{IEEEtran}
\bibliography{ref}

\begin{thebibliography}{10}
\providecommand{\url}[1]{#1}
\csname url@samestyle\endcsname
\providecommand{\newblock}{\relax}
\providecommand{\bibinfo}[2]{#2}
\providecommand{\BIBentrySTDinterwordspacing}{\spaceskip=0pt\relax}
\providecommand{\BIBentryALTinterwordstretchfactor}{4}
\providecommand{\BIBentryALTinterwordspacing}{\spaceskip=\fontdimen2\font plus
\BIBentryALTinterwordstretchfactor\fontdimen3\font minus
  \fontdimen4\font\relax}
\providecommand{\BIBforeignlanguage}[2]{{%
\expandafter\ifx\csname l@#1\endcsname\relax
\typeout{** WARNING: IEEEtran.bst: No hyphenation pattern has been}%
\typeout{** loaded for the language `#1'. Using the pattern for}%
\typeout{** the default language instead.}%
\else
\language=\csname l@#1\endcsname
\fi
#2}}
\providecommand{\BIBdecl}{\relax}
\BIBdecl

\bibitem{r2}
B.~Oberfeld, A.~Achanta, K.~Carpenter, P.~Chen, N.~M. Gilette, P.~Langat, J.~T.
  Said, A.~E. Schiff, A.~S. Zhou, A.~K. Barczak \emph{et~al.}, ``{S}nap{S}hot:
  covid-19,'' \emph{Cell}, vol. 181, no.~4, pp. 954--954, 2020.

\bibitem{r3}
V.~A. Rodriguez, S.~Bhave, R.~Chen, C.~Pang, G.~Hripcsak, S.~Sengupta,
  N.~Elhadad, R.~Green, J.~Adelman, K.~S. Metitiri \emph{et~al.}, ``Development
  and validation of prediction models for mechanical ventilation, renal
  replacement therapy, and readmission in {COVID}-19 patients,'' \emph{Journal
  of the American Medical Informatics Association}, vol.~28, no.~7, pp.
  1480--1488, 2021.

\bibitem{peiris2022hospital}
S.~Peiris, J.~L. Nates, J.~Toledo, Y.-L. Ho, O.~Sosa, V.~Stanford,
  S.~Aldighieri, and L.~Reveiz, ``Hospital readmissions and emergency
  department re-presentation of {COVID}-19 patients: a systematic review,''
  \emph{Rev Panam Salud Publica; 46, oct. 2022}, 2022.

\bibitem{pan2010survey}
S.~J. Pan and Q.~Yang, ``A survey on transfer learning,'' \emph{IEEE
  Transactions on knowledge and data engineering}, vol.~22, no.~10, pp.
  1345--1359, 2010.

\bibitem{zhao2020review}
S.~Zhao, X.~Yue, S.~Zhang, B.~Li, H.~Zhao, B.~Wu, R.~Krishna, J.~E. Gonzalez,
  A.~L. Sangiovanni-Vincentelli, S.~A. Seshia \emph{et~al.}, ``A review of
  single-source deep unsupervised visual domain adaptation,'' \emph{IEEE
  Transactions on Neural Networks and Learning Systems}, vol.~33, no.~2, pp.
  473--493, 2020.

\bibitem{ruder2019transfer}
S.~Ruder, M.~E. Peters, S.~Swayamdipta, and T.~Wolf, ``Transfer learning in
  natural language processing,'' in \emph{Proceedings of the 2019 conference of
  the North American chapter of the association for computational linguistics:
  Tutorials}, 2019, pp. 15--18.

\bibitem{balachandar2020accounting}
N.~Balachandar, K.~Chang, J.~Kalpathy-Cramer, and D.~L. Rubin, ``Accounting for
  data variability in multi-institutional distributed deep learning for medical
  imaging,'' \emph{Journal of the American Medical Informatics Association},
  vol.~27, no.~5, pp. 700--708, 2020.

\bibitem{muhammad2020deep}
K.~Muhammad, S.~Khan, J.~Del~Ser, and V.~H.~C. De~Albuquerque, ``Deep learning
  for multigrade brain tumor classification in smart healthcare systems: A
  prospective survey,'' \emph{IEEE Transactions on Neural Networks and Learning
  Systems}, vol.~32, no.~2, pp. 507--522, 2020.

\bibitem{alzubaidi2021role}
M.~Alzubaidi, H.~D. Zubaydi, A.~A. Bin-Salem, A.~A. Abd-Alrazaq, A.~Ahmed, and
  M.~Househ, ``Role of deep learning in early detection of {COVID}-19: Scoping
  review,'' \emph{Computer methods and programs in biomedicine update}, vol.~1,
  p. 100025, 2021.

\bibitem{zhao2020multi}
Z.~Zhao, J.~Qin, Z.~Gou, Y.~Zhang, and Y.~Yang, ``Multi-task learning models
  for predicting active compounds,'' \emph{Journal of Biomedical Informatics},
  vol. 108, p. 103484, 2020.

\bibitem{kim2021anticancer}
Y.~Kim, S.~Zheng, J.~Tang, W.~Jim~Zheng, Z.~Li, and X.~Jiang, ``Anticancer drug
  synergy prediction in understudied tissues using transfer learning,''
  \emph{Journal of the American Medical Informatics Association}, vol.~28,
  no.~1, pp. 42--51, 2021.

\bibitem{sachan2018effective}
D.~S. Sachan, P.~Xie, M.~Sachan, and E.~P. Xing, ``Effective use of
  bidirectional language modeling for transfer learning in biomedical named
  entity recognition,'' in \emph{Machine learning for healthcare
  conference}.\hskip 1em plus 0.5em minus 0.4em\relax PMLR, 2018, pp. 383--402.

\bibitem{lin2020does}
C.~Lin, S.~Bethard, D.~Dligach, F.~Sadeque, G.~Savova, and T.~A. Miller, ``Does
  {BERT} need domain adaptation for clinical negation detection?''
  \emph{Journal of the American Medical Informatics Association}, vol.~27,
  no.~4, pp. 584--591, 2020.

\bibitem{dana2018deep}
D.~Dana, S.~V. Gadhiya, L.~G. St.~Surin, D.~Li, F.~Naaz, Q.~Ali, L.~Paka, M.~A.
  Yamin, M.~Narayan, I.~D. Goldberg \emph{et~al.}, ``Deep learning in drug
  discovery and medicine; scratching the surface,'' \emph{Molecules}, vol.~23,
  no.~9, p. 2384, 2018.

\bibitem{saeedi2020collaborative}
R.~Saeedi, K.~Sasani, and A.~H. Gebremedhin, ``Collaborative multi-expert
  active learning for mobile health monitoring: architecture, algorithms, and
  evaluation,'' \emph{Sensors}, vol.~20, no.~7, p. 1932, 2020.

\bibitem{ganin2016domain}
Y.~Ganin, E.~Ustinova, H.~Ajakan, P.~Germain, H.~Larochelle, F.~Laviolette,
  M.~Marchand, and V.~Lempitsky, ``Domain-adversarial training of neural
  networks,'' \emph{The journal of machine learning research}, vol.~17, no.~1,
  pp. 2096--2030, 2016.

\bibitem{brion2021domain}
E.~Brion, J.~L{\'e}ger, A.~M. Barrag{\'a}n-Montero, N.~Meert, J.~A. Lee, and
  B.~Macq, ``Domain adversarial networks and intensity-based data augmentation
  for male pelvic organ segmentation in cone beam {CT},'' \emph{Computers in
  Biology and Medicine}, vol. 131, p. 104269, 2021.

\bibitem{lafarge2017domain}
M.~W. Lafarge, J.~P. Pluim, K.~A. Eppenhof, P.~Moeskops, and M.~Veta,
  ``Domain-adversarial neural networks to address the appearance variability of
  histopathology images,'' in \emph{Deep Learning in Medical Image Analysis and
  Multimodal Learning for Clinical Decision Support: Third International
  Workshop, DLMIA 2017, and 7th International Workshop, ML-CDS 2017, Held in
  Conjunction with MICCAI 2017, Qu{\'e}bec City, QC, Canada, September 14,
  Proceedings 3}.\hskip 1em plus 0.5em minus 0.4em\relax Springer, 2017, pp.
  83--91.

\bibitem{ye2021deep}
Y.~Ye and A.~Gu, ``Deep transfer learning for infectious disease case detection
  using electronic medical records,'' \emph{arXiv preprint arXiv:2103.06710},
  2021.

\bibitem{r20}
S.~Visweswaran, B.~McLay, N.~Cappella, M.~Morris, J.~T. Milnes, S.~E. Reis,
  J.~C. Silverstein, and M.~J. Becich, ``An atomic approach to the design and
  implementation of a research data warehouse,'' \emph{Journal of the American
  Medical Informatics Association}, vol.~29, no.~4, pp. 601--608, 2022.

\bibitem{r21}
M.~Neumann, D.~King, I.~Beltagy, and W.~Ammar, ``{S}cispa{C}y: Fast and robust
  models for biomedical natural language processing,'' in \emph{Proceedings of
  the 18th BioNLP Workshop and Shared Task}.\hskip 1em plus 0.5em minus
  0.4em\relax Association for Computational Linguistics, 2019, pp. 319--327.

\bibitem{day2017survey}
O.~Day and T.~M. Khoshgoftaar, ``A survey on heterogeneous transfer learning,''
  \emph{Journal of Big Data}, vol.~4, pp. 1--42, 2017.

\bibitem{zhuang2020comprehensive}
F.~Zhuang, Z.~Qi, K.~Duan, D.~Xi, Y.~Zhu, H.~Zhu, H.~Xiong, and Q.~He, ``A
  comprehensive survey on transfer learning,'' \emph{Proceedings of the IEEE},
  vol. 109, no.~1, pp. 43--76, 2020.

\bibitem{jiang2022transferability}
J.~Jiang, Y.~Shu, J.~Wang, and M.~Long, ``Transferability in deep learning: A
  survey,'' \emph{arXiv preprint arXiv:2201.05867}, 2022.

\bibitem{r22}
S.~Sun, H.~Shi, and Y.~Wu, ``A survey of multi-source domain adaptation,''
  \emph{Information Fusion}, vol.~24, pp. 84--92, 2015.

\bibitem{r23}
K.~Li, J.~Lu, H.~Zuo, and G.~Zhang, ``Multi-source contribution learning for
  domain adaptation,'' \emph{IEEE Transactions on Neural Networks and Learning
  Systems}, vol.~33, no.~10, pp. 5293--5307, 2021.

\bibitem{r24}
S.~Christodoulidis, M.~Anthimopoulos, L.~Ebner, A.~Christe, and S.~Mougiakakou,
  ``Multisource transfer learning with convolutional neural networks for lung
  pattern analysis,'' \emph{IEEE journal of biomedical and health informatics},
  vol.~21, no.~1, pp. 76--84, 2016.

\bibitem{bengio2013representation}
Y.~Bengio, A.~Courville, and P.~Vincent, ``Representation learning: A review
  and new perspectives,'' \emph{IEEE transactions on pattern analysis and
  machine intelligence}, vol.~35, no.~8, pp. 1798--1828, 2013.

\bibitem{dou2022learning}
J.~X. Dou, M.~Jia, N.~Zaslavsky, M.~Ebeid, R.~Bao, S.~Zhang, K.~Ni, P.~P.
  Liang, H.~Mao, and Z.-H. Mao, ``Learning more effective cell representations
  efficiently,'' in \emph{NeurIPS 2022 Workshop on Learning Meaningful
  Representations of Life}, 2022.

\bibitem{dou2022sampling}
J.~X. Dou, A.~Q. Pan, R.~Bao, H.~H. Mao, and L.~Luo, ``Sampling through the
  lens of sequential decision making,'' \emph{arXiv preprint arXiv:2208.08056},
  2022.

\bibitem{alsentzer2019publicly}
E.~Alsentzer, J.~R. Murphy, W.~Boag, W.-H. Weng, D.~Jin, T.~Naumann, and
  M.~McDermott, ``Publicly available clinical {BERT} embeddings,'' \emph{arXiv
  preprint arXiv:1904.03323}, 2019.

\bibitem{luo2022biogpt}
R.~Luo, L.~Sun, Y.~Xia, T.~Qin, S.~Zhang, H.~Poon, and T.-Y. Liu, ``Bio{GPT}:
  generative pre-trained transformer for biomedical text generation and
  mining,'' \emph{Briefings in Bioinformatics}, vol.~23, no.~6, 2022.

\bibitem{chatgpt}
\BIBentryALTinterwordspacing
OpenAI, ``Introducing chat{GPT}.'' [Online]. Available:
  \url{https://openai.com/blog/chatgpt}
\BIBentrySTDinterwordspacing

\bibitem{kung2023performance}
T.~H. Kung, M.~Cheatham, A.~Medenilla, C.~Sillos, L.~De~Leon, C.~Elepa{\~n}o,
  M.~Madriaga, R.~Aggabao, G.~Diaz-Candido, J.~Maningo \emph{et~al.},
  ``Performance of {chatGPT} on {USMLE}: Potential for {AI} -assisted medical
  education using large language models,'' \emph{PLoS digital health}, vol.~2,
  no.~2, p. e0000198, 2023.

\end{thebibliography}

\vspace{12pt}

\end{document}